\documentclass[runningheads]{llncs}
\usepackage[T1]{fontenc}
\usepackage{graphicx}
\usepackage{booktabs}
\usepackage{amsmath,amssymb}
\usepackage{xcolor}
\usepackage{url}
\usepackage{float}
\usepackage{subcaption}
\usepackage[hidelinks]{hyperref}

\begin{document}
\title{Parameter-Efficient pretrained-CT-to-MRI Transfer for Rectal Cancer Segmentation: Performance-Calibration Trade-offs}
%
\titlerunning{Parameter-Efficient CT-to-MRI Transfer for Rectal Cancer}
%
\author{Aneesh Rangnekar\orcidID{0000-0002-0079-9495} \and
Jorge Tapias Gomez \and
Joseph O Deasy \and
Harini Veeraraghavan}
\authorrunning{A. Rangnekar et al.}
\institute{Memorial Sloan Kettering Cancer Center, New York, USA
\\ 
Corresponding author: \email{veerarah@mskcc.org}
}

\maketitle              
\begin{abstract}
Accurate rectal cancer segmentation from magnetic resonance imaging (MRI) is essential for adaptive radiotherapy and tumor response assessment, but deployment also requires computational efficiency and informative, calibrated uncertainty estimates. We therefore introduce SWIFT, a \textbf{SW}in pretrained model w\textbf{i}th parameter-e\textbf{F}ficient and \textbf{T}umor-aware fine-tuning for rectal cancer segmentation. A Swin~V2 encoder pretrained on 10,444 public 3D CT volumes using a DINOv2-style objective was adapted to T2-weighted MRI through four cumulative configurations: full fine-tuning (SWIFT), decoder compression (SWIFTe), low-rank adaptation (SWIFTe-LoRA), and a four-member LoRA-decoder ensemble (SWIFTe-LDE4). Geometric accuracy, tumor detection, radiomic agreement, and probability calibration were evaluated on a held-out 247-case test set from a single-institution cohort acquired using 1.5 or 3~Tesla GE scanners. Compared with SWIFT, SWIFTe reduced total parameters by 70.1\% (from 72.8M to 21.8M) and increased tumor detection rate from 89.9\% to 93.9\%, while achieving a slightly lower median surface DSC (0.61 versus 0.62) and improved radiomic agreement. In a separate SWIFTe ablation, removing tumor-aware augmentation reduced detection from 93.9\% to 89.9\% but increased surface DSC from 0.61 to 0.64, demonstrating a detection-boundary-agreement trade-off. SWIFTe-LoRA used 14.6\% of SWIFTe's trainable parameters while retaining similar segmentation performance. SWIFTe-LDE4 achieved the lowest calibration errors among the four configurations after temperature scaling (expected calibration error, 0.217; Brier score, 0.222), although the absolute expected calibration error indicates residual miscalibration. Similar efficiency-calibration patterns were observed using the public VoCo checkpoint, supporting robustness across pretrained initializations rather than external clinical generalizability. Code and selected model checkpoints are available in our \href{https://github.com/The-Veeraraghavan-Lab/RectalTxSeg}{\textcolor{blue}{GitHub repository}}.

\end{abstract}

\section{Introduction}
\label{sec:introduction:motivation}

MR-guided adaptive radiotherapy (MRgART) enables daily treatment adaptation to deliver ablative tumor doses while limiting exposure to adjacent radiosensitive gastrointestinal organs~\cite{Henke2016SimulatedOnlineAdaptive}. Daily rectal tumor contouring is a major bottleneck in MRgART~\cite{Intven2021OnlineAdaptiveRectal}. Self-supervised training of high-capacity deep learning (DL) models with diverse imaging datasets has demonstrated adaptability to multiple medical imaging tasks~\cite{tang2022self,jiang2022self,goyal2021self,wu2025large} and robustness to some imaging variations~\cite{jiang2024self}. However, models can fail in clinical scenarios not captured during training, requiring manual contouring~\cite{Kensen2024PatientSpecificRectal}. Uncertainty metrics generated from voxel-wise probabilistic predictions can provide information about segmentation confidence~\cite{Ren2024EnhancingReliability,Rangnekar2025QuantifyingUncertainty} and potentially guide the manual correction process~\cite{Wei2025InteractiveDeepLearning,Maruccio2025NetworkUncertaintyRectal}. However, poorly calibrated models may fail confidently in unfamiliar testing regimes~\cite{mehrtash2020confidence,karimi2022improving,rangnekar2026tumoranchored}. Furthermore, large models require significant computational resources for adaptation, inference, and uncertainty estimation, creating a computational bottleneck in a complex adaptive radiotherapy (aRT) workflow. 

Published rectal-tumor MRI studies reported DSC values of 0.68--0.70 on multiparametric MRI~\cite{trebeschi2017deep}, while patient-specific adaptation on daily MR-Linac images reduced median HD95 from 22.0~mm to 4.8~mm~\cite{Kensen2024PatientSpecificRectal}. Differences in cohorts, protocols, and metrics preclude direct ranking against our results.

DL models deployed in time-sensitive aRT settings must often satisfy three requirements: (i) acceptable geometric accuracy, (ii) practical computational costs, and (iii) calibrated uncertainty estimates to guide when user intervention is needed. Prior works have individually examined methods to improve geometric accuracy~\cite{Kensen2024PatientSpecificRectal}, reliability of uncertainties~\cite{Ren2024EnhancingReliability,rangnekar2026tumoranchored}, and strategies to address computational bottlenecks, including streamlined decoders~\cite{rahman2025effidec3d}, low-rank adaptation (LoRA) for efficient tuning~\cite{hu2022lora}, and probabilistic inference with shared heads and LoRA~\cite{fuchs2021practical,abboud2024sparse,halbheer2025loraensemble}. However, none have studied the interplay of parameter efficiency, performance, and calibration for reliably accurate tumor segmentation considering clinical deployment. Our key contribution is therefore a benchmarking framework that holistically evaluates performance-calibration trade-offs with parameter efficiency in pretrained models.

We benchmarked a cumulative four-stage curriculum by first creating a model called \textbf{S}win pretrained model \textbf{w}ith parameter e\textbf{F}ficient and \textbf{T}umor-aware fine-tuning (SWIFT) applied to rectal cancer segmentation from MRI. SWIFT uses Swin transformer~V2 to extract computationally efficient attention, integrating local and global anatomical context while supporting stable training for improved transferability~\cite{he2023swinunetr}. It retains the standard Swin~UNETR~V2 decoder~\cite{he2023swinunetr} and is fine-tuned end-to-end, providing a full-adaptation reference. Besides imaging differences, rectal cancers exhibit a highly variable appearance on T2-weighted MRI due to tumor morphology, histologic composition, lymph node invasion, and extracellular mucin, confounding accurate segmentation~\cite{Horvat2019MRIRectalCancer,battersby2016prospective,lin2023fully,yang202421}. To address this, we developed a tumor-aware intensity augmentation that selectively varies the intensity statistics within tumors to capture intermediate, dark, and bright tumors. All model variants were fine-tuned with tumor-aware augmentation. Second, we evaluated the impact of reducing decoder capacity and model complexity by replacing the original decoder with an efficient EffiDec3D decoder~\cite{rahman2025effidec3d} within SWIFT (or SWIFTe). Third, we analyzed the impact of freezing the pretrained encoders to reduce the parameters for fine-tuning using LoRA~\cite{hu2022lora} within SWIFTe (or SWIFTe-LoRA). Fourth, we evaluated whether extending SWIFTe-LoRA with member-specific adapter-decoder pairs and shared pretrained encoder (or SWIFTe-LDE4) enhanced uncertainty calibration~\cite{halbheer2025loraensemble}.

\section{Methodology}
\label{sec:methods}

\begin{figure}[t]
\centering
\begin{minipage}[c]{0.42\textwidth}
\centering
\includegraphics[width=\textwidth]{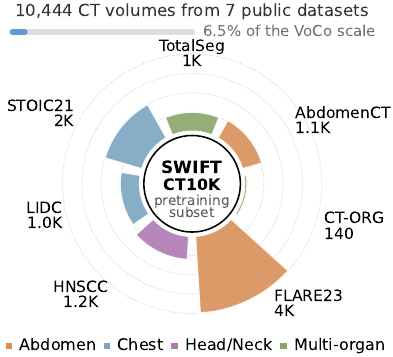}
\end{minipage}
\hfill
\begin{minipage}[c]{0.56\textwidth}
\centering
\begin{tabular}{@{}cc@{}}
\includegraphics[width=0.48\textwidth]{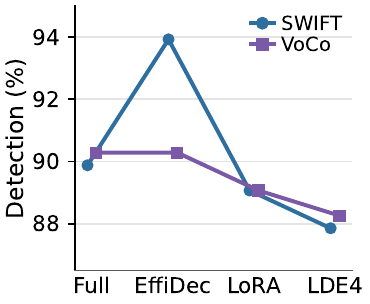} &
\includegraphics[width=0.48\textwidth]{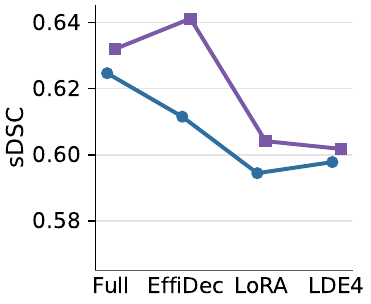} \\[-0.2ex]
\includegraphics[width=0.48\textwidth]{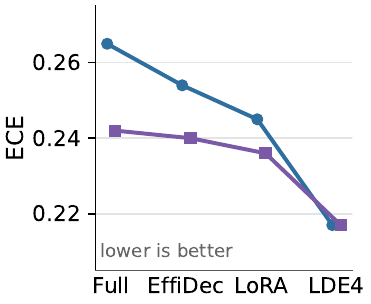} &
\includegraphics[width=0.48\textwidth]{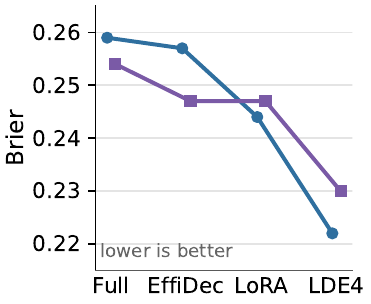}
\end{tabular}
\end{minipage}
\caption{Pretraining context and transfer results. Left: CT10K pretraining subset used for SWIFT encoder pretraining and its scale relative to VoCo. Right: detection, sDSC, ECE, and Brier score across full fine-tuning, efficient decoder (EffiDec), LoRA adaptation, and LDE4 for SWIFT and VoCo.}
\label{fig:transfer_summary}
\end{figure}

\begin{figure}[t]
\centering
\includegraphics[width=0.98\textwidth]{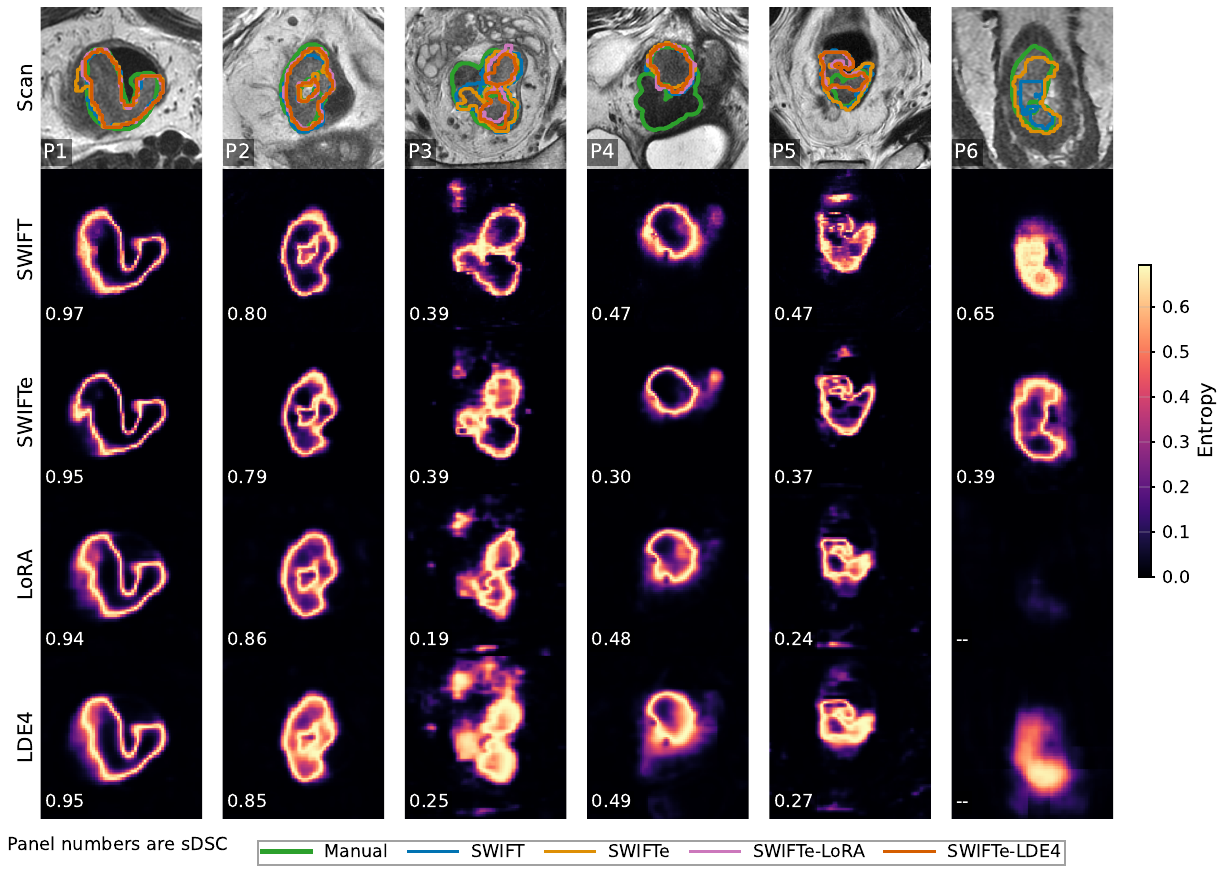}
\caption{Representative predictive-entropy maps across six test cases. The top row shows the image with manual and automated contours; remaining rows show entropy maps for SWIFT, SWIFTe, SWIFTe-LoRA, and SWIFTe-LDE4. Yellow indicates higher entropy, and panel numbers denote sDSC. }
\label{fig:uncertainty_maps}
\end{figure}

\textbf{Data preprocessing:}
This retrospective study involving patients with rectal cancers imaged before treatment was performed following institutional review board approval and waiver of consent. A total of 416 T2-weighted oblique axial MRIs acquired on 1.5 or 3 Tesla GE scanners with phased-array coils, 2 to 4~mm slice thickness, and field of view $\sim$180 to 220~mm were analyzed using patient-specific splits for training (n = 136), validation (n = 33), and testing (n = 247). Images were resampled to 1~mm$^3$ isotropic spacing, reoriented to RAS, and clipped to $[0,800]$. Network inputs were min-max scaled after clipping, whereas radiomics used the clipped intensity scale to preserve the engineered-feature convention. Manual tumor contours served as the reference standard.
\\
\\
\textbf{Full fine-tuning baseline (SWIFT):}
SWIFT is based on a CT-to-MRI cross-modality transfer pipeline: a 3D Swin~UNETR~V2~\cite{he2023swinunetr} network is first pre-trained in a self-supervised manner on 10,444 heterogeneous public CT volumes (Fig.~\ref{fig:transfer_summary}) via a DINOv2-style~\cite{oquab2024dinov} self-distillation objective, then fine-tuned for segmenting rectal cancers on MRI. The model consists of a hierarchical Swin~V2 encoder and a Swin~UNETR decoder with multi-scale skip connections, using a feature size of 48 and a two-channel background/tumor output. In SWIFT, the full encoder-decoder network is fine-tuned end-to-end, serving as the baseline architecture for benchmarking remaining configurations as described below.
\\
\\
\textbf{Efficient decoder:}
SWIFTe modifies SWIFT by replacing the Swin~UNETR decoder with EffiDec3D~\cite{rahman2025effidec3d}, while retaining the CT pretrained Swin~V2 encoder and downstream training/inference protocol, allowing us to isolate the effect of decoder compression. EffiDec3D uses a fixed channel width across stages (48 decoder channels) and a resolution factor of 2. Omitting the finest full-resolution reconstruction stage, it reconstructs logits at reduced spatial resolution before trilinear upsampling to the input grid.
\\
\\
\textbf{SWIFTe-LoRA and SWIFTe-LDE4:}
We next tested encoder-efficient adaptation with SWIFTe-LoRA and probability calibration with SWIFTe-LDE4. SWIFTe-LoRA used parameter-efficient adaptation~\cite{hu2022lora} of SWIFTe by inserting LoRA adapters into the Swin window-attention query, key, value, and output-projection layers after loading the CT-pretrained encoder. The frozen pretrained encoder was adapted using trainable rank-4 LoRA modules, while jointly training the EffiDec3D decoder. SWIFTe-LDE4 extended this setting to four member-specific LoRA-adapter and decoder pairs~\cite{halbheer2025loraensemble} with a shared frozen pretrained encoder. We used four ensemble members as a pragmatic compromise between ensemble diversity and computational cost, since each member adds an adapter-decoder branch and forward pass. Foreground probabilities were averaged for segmentation, while inter-member variance and disagreement quantified uncertainty.
\\
\\
\textbf{Implementation details:}
All configurations were optimized with a combined soft Dice and cross-entropy loss using AdamW, a learning rate of $3\times10^{-4}$, weight decay $10^{-5}$, linear warmup, and cosine annealing. During training only, tumor-aware intensity augmentation used the manual tumor mask to simulate rectal-tumor appearance variation by locally transforming foreground tumor voxels as $I'(x)=\alpha I(x)+\beta$, where $\alpha\sim\mathcal{U}(0.5,1.5)$ and $\beta\sim\mathcal{U}(-0.2,0.2)$. Surrounding anatomy was left unchanged, and the transform was applied after spatial augmentation with probability 0.3. LoRA models used the same loss, optimizer family, crop size, augmentation, and inference protocol, with optimization restricted to the trainable adapter and decoder parameters described above. Tumor-aware augmentation was held constant across the four primary configurations and was not part of the parameter-reduction strategy.
\\
The models used 3D crops of $96\times96\times64$, instance normalization, and a two-class foreground/background output. Binary masks were generated using a foreground probability threshold of $\tau=0.5$. Inference used 3D sliding windows with 50\% overlap and test-time augmentation. Automated postprocessing selected the largest connected component as tumor corresponding to the primary gross tumor volume. Inference operated on full volumes without manual-contour information. Detection included all test cases, while model-specific misses were penalized in the primary sDSC analysis as described below.
\\
\\
\textbf{Case-level uncertainty:}
For all models, case-level uncertainty was summarized as mean predictive entropy within the predicted tumor. Empty predictions for missed cases were represented by averaging the uncertainty map over the full evaluation volume. For SWIFTe-LDE4, member variance of tumor probability and member disagreement were additionally summarized as ensemble-diversity measures. Retrospective calibration evaluation used a tumor-focused region of interest comprising voxels with tumor probability $\geq0.05$ or manual tumor label. Representative entropy maps are shown in Fig.~\ref{fig:uncertainty_maps}.

\textbf{Evaluation metrics:}
Segmentation performance across 247 test cases was evaluated using detection rate (prediction-reference intersection-over-union $>0.1$) and surface DSC (sDSC) at a 2~mm tolerance. HD95 was not reported because it is undefined for empty predictions, and detected-case-only reporting would produce different model-specific cohorts that could favor models with more misses. We therefore report detection over all cases and assign zero sDSC to model-specific misses within the shared evaluable cohort. Primary sDSC summaries analyzed a shared-detectable cohort of 236 cases, setting model-specific misses to zero and excluding 11 cases missed by all evaluated models, while Lin’s concordance correlation coefficient evaluated agreement across IBSI-aligned radiomic feature families. Paired inferential comparisons applied Cochran’s Q and exact McNemar tests for detection alongside Friedman and Wilcoxon signed-rank tests for sDSC, with $p$-values adjusted using the Holm procedure. Reliability diagrams, expected calibration error (ECE), and Brier score were reported following validation-set temperature scaling. 

\section{Results}
\label{sec:results:overview}

\textbf{SWIFTe achieved higher detection with fewer parameters:}
SWIFTe reduced total parameters by 70.1\%, from 72.8M for SWIFT to 21.8M, while detecting 232 of 247 tumors (93.9\%) and retaining a shared-detectable median sDSC of 0.61 (Fig.~\ref{fig:transfer_summary}). Although SWIFT reached a slightly higher median sDSC of 0.62, its detection rate was 89.9\% compared with 93.9\% for SWIFTe; detection rates differed across configurations (Cochran's Q, Holm-adjusted $p=0.039$).
\\
\\
\textbf{SWIFTe-LDE4 improved calibration while retaining similar sDSC:}
SWIFTe-LDE4 achieved the lowest calibration errors among the SWIFT configurations, reducing ECE from 0.245 for SWIFTe-LoRA to 0.217 and Brier score from 0.244 to 0.222 (Fig.~\ref{fig:transfer_summary}). Median sDSC remained similar between SWIFTe-LDE4 and SWIFTe-LoRA (0.60 versus 0.59), while detection was 87.9\% and 89.1\%, respectively. Probability-derived entropy was available for all configurations, while SWIFTe-LDE4 additionally provided member-variance and disagreement measures for potential review prioritization. Although lowest among the evaluated configurations, an ECE of 0.217 indicates residual absolute miscalibration.

\begin{figure}[t]
    \centering

    \begin{subfigure}[t]{0.24\textwidth}
        \centering
        \includegraphics[width=\linewidth]{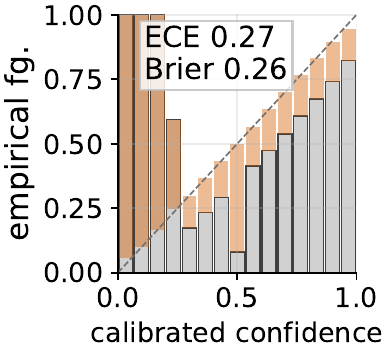}
        \caption{SWIFT.}
        \label{fig:calibration_swift}
    \end{subfigure}\hfill
    \begin{subfigure}[t]{0.24\textwidth}
        \centering
        \includegraphics[width=\linewidth]{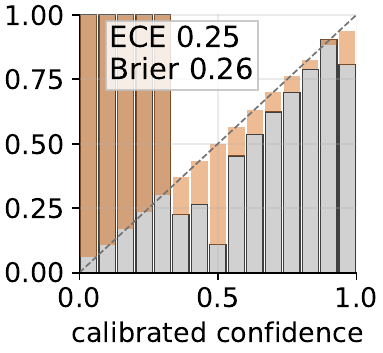}
        \caption{SWIFTe.}
        \label{fig:calibration_swifte}
    \end{subfigure}\hfill
    \begin{subfigure}[t]{0.24\textwidth}
        \centering
        \includegraphics[width=\linewidth]{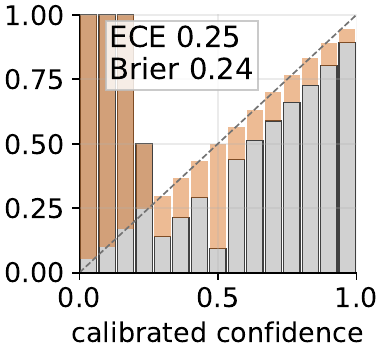}
        \caption{SWIFTe-LoRA.}
        \label{fig:calibration_swifte_lora}
    \end{subfigure}\hfill
    \begin{subfigure}[t]{0.24\textwidth}
        \centering
        \includegraphics[width=\linewidth]{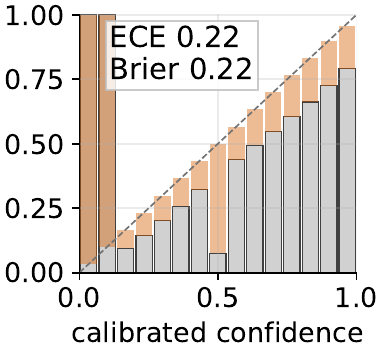}
        \caption{SWIFTe-LDE4.}
        \label{fig:calibration_swifte_lde4}
    \end{subfigure}

    \caption{Reliability diagrams show foreground probability calibration for each SWIFT configuration. Gray bars show empirical foreground frequency, orange shading shows the calibration gap, and the dashed line denotes ideal calibration. SWIFTe-LDE4 achieved the lowest ECE and Brier score, illustrating that calibration and segmentation overlap are distinct objectives.}
    \label{fig:lora_uncertainty}
\end{figure}

\begin{figure}[t]
\centering
\includegraphics[width=0.92\textwidth]{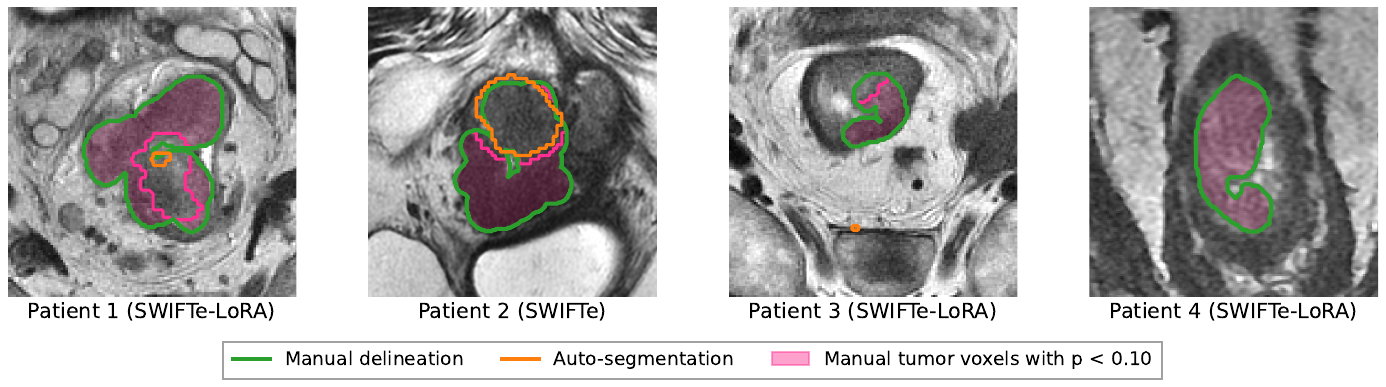}
\caption{Low-confidence tumor examples from calibration analysis. Pink regions indicate manual tumor voxels assigned foreground probability $p<0.10$, with manual contours in green and automated contours in orange.}
\label{fig:low_confidence_examples}
\end{figure}

One scalar temperature per configuration was fitted on the 33-case validation set and applied unchanged to the test set (Fig.~\ref{fig:lora_uncertainty}). Because temperature scaling preserved the $0.5$ decision boundary, thresholded contours and segmentation metrics were unchanged. Low-confidence calibration regions corresponded qualitatively to tumor voxels assigned low foreground probability in challenging cases typically consisting of tumors with heterogeneous T2 signal and diffuse boundaries (Fig.~\ref{fig:low_confidence_examples}).

\begin{figure}[t]
    \centering

    \begin{subfigure}[t]{0.58\textwidth}
        \centering
        \includegraphics[width=\linewidth]{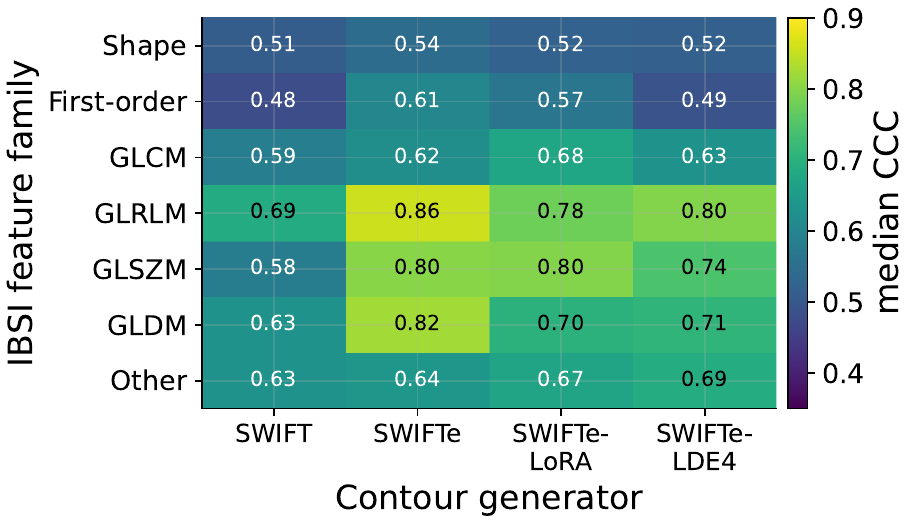}
        \caption{Median CCC by IBSI feature family.}
        \label{fig:radiomics_check_family}
    \end{subfigure}
    \hspace{0.02\textwidth}
    \begin{subfigure}[t]{0.33\textwidth}
        \centering
        \includegraphics[width=\linewidth]{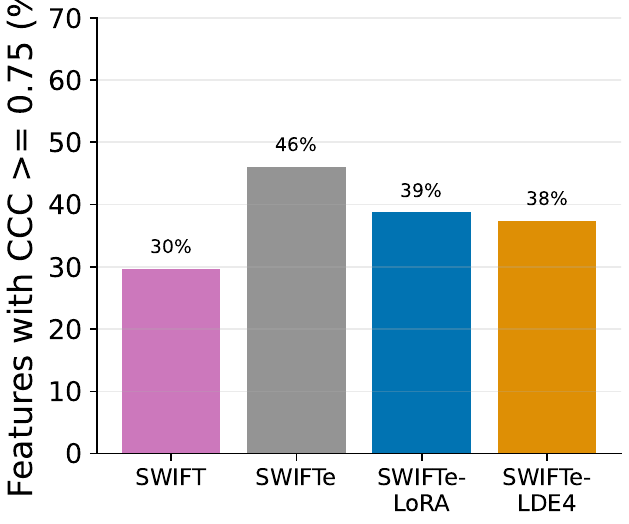}
        \caption{Fraction of features with CCC $\geq 0.75$.}
        \label{fig:radiomics_check_fraction}
    \end{subfigure}

    \caption{Radiomic-feature agreement across the SWIFT progression. Panel~(a) shows median manual-versus-automated CCC by IBSI feature family; panel~(b) shows the fraction of features with CCC $\geq 0.75$.}
    \label{fig:radiomics_check}
\end{figure}

\noindent\textbf{Several radiomic feature families showed strong agreement:}
Radiomic agreement varied across feature families and SWIFT configurations (Fig.~\ref{fig:radiomics_check}). Across 247 matched cases and 288 numeric features, median manual-versus-automated CCC was 0.62 for SWIFT, 0.68 for SWIFTe, and 0.69 for both SWIFTe-LoRA and SWIFTe-LDE4. The fraction of features with CCC $\geq0.75$ was 30\%, 46\%, 39\%, and 38\%, respectively. Shape features were the least reproducible family across all configurations (median CCC 0.51--0.54), consistent with modest median sDSC of 0.6. SWIFTe considerably enhanced reproducibility of first-order histogram and intra-tumor texture heterogeneity measures, including the GLRLM, GLSZM, and GLDM features. The LoRA-adapted configuration also improved reproducibility of intra-tumor texture heterogeneity feature families, including GLCM and GLSZM, compared with full fine-tuning.
\\
\\
\textbf{LoRA retained similar sDSC with fewer trainable parameters:}
SWIFTe-LoRA reduced the trainable burden to 3.2M parameters, corresponding to 14.6\% of SWIFTe, while retaining a similar median sDSC (0.59 versus 0.61). Detection rate was lower at 89.1\% for SWIFTe-LoRA compared to SWIFTe (93.9\%). SWIFTe-LDE4 used 12.7M trainable parameters across four member-specific adapter-decoder pairs and similarly retained median sDSC at 0.60, while providing improved probability calibration.
\\
\\
\textbf{The efficiency-calibration pattern was reproducible with VoCo:}
We replicated the decoder and adaptation sequence using VoCo-pretrained weights to assess sensitivity to initialization with a model pretrained using a substantially larger dataset (Fig.~\ref{fig:transfer_summary}). Because SWIFT and VoCo differ in pretraining scale and objective, this was not a direct comparison of pretraining methods, but a test of whether the efficiency–segmentation–calibration trends generalized across initializations. The VoCo progression showed the same overall pattern: efficient decoding reduced model size while preserving segmentation performance, LoRA reduced the trainable burden while retaining similar overlap, and LDE4 improved probability calibration while maintaining broadly similar sDSC.
\\
\\
\textbf{Ablation study:} Removing tumor-aware intensity augmentation reduced detection for SWIFTe (93.9\% to 89.9\%) while increasing sDSC from 0.61 to 0.64. SWIFT showed the same detection trend. The transform was retained to prioritize detection over boundary agreement.

For ensemble sizing, we evaluated SWIFTe-LDE variants with 2, 4, and 8 members. Four members provided a pragmatic balance between performance and computational cost, as larger ensembles offered minimal performance gains for added compute.

Temperature scaling reduced Brier scores across all configurations and lowered ECE for SWIFT, SWIFTe, and SWIFTe-LoRA without altering thresholded segmentations. Although SWIFTe-LDE4 ECE rose slightly from 0.207 to 0.217, it remained the lowest among all evaluated configurations.

\section{Discussion and conclusions}
\label{sec:discussion:scope}

This study benchmarked a cumulative parameter-efficient cross-modality adaptation approach using a CT-pretrained Swin transformer for rectal cancer segmentation from MRI. We found that an efficient decoder that reduced total parameters by 70.1\% maintained segmentation performance while enhancing detection performance. LoRA adapter configurations achieved detection performance comparable to full fine-tuning, while the SWIFTe-LDE4 configuration improved calibration relative to the other SWIFT configurations. Importantly, VoCo showed the same behavior with parameter-efficient adaptation even when pretrained using substantially more examples. These results support efficient decoding and LoRA as parameter-efficient adaptation strategies, with LDE4 providing a relative calibration improvement without substantially degrading segmentation accuracy.

The downstream analyses further showed why overlap alone was insufficient to characterize model behavior. Radiomic agreement varied by feature family: texture features, particularly those reflecting intra-tumor heterogeneity, were more reproducible for SWIFTe and the LoRA-adapted configurations, whereas shape features remained less reproducible across all configurations, consistent with the modest geometric agreement against manual contours. The models also showed systematic contour-volume differences, indicating that similar median sDSC can still correspond to different boundary behavior. Finally, higher entropy and, for SWIFTe-LDE4, higher ensemble disagreement generally corresponded to lower-confidence cases, supporting their use as qualitative review-prioritization signals rather than validated spatial error-localization maps. 

This study evaluated one tumor type using a single-institution cohort acquired exclusively on GE scanners; robustness across institutions, vendors, and imaging protocols remains unverified. Comparisons were primarily internal to the SWIFT family, and published results provide context rather than head-to-head benchmarks. The uncertainty signals were descriptive and were not quantitatively validated against manual correction effort or dosimetric impact. Multi-institutional validation and prospective evaluation of uncertainty-guided review are needed before clinical deployment.

Within the present retrospective cohort, parameter-efficient CT-pretrained transformer adaptations preserved segmentation performance, while SWIFTe-LDE4 improved calibration relative to the other evaluated configurations. Further validation of efficient uncertainty-aware segmentation workflows for adaptive radiotherapy is warranted.

\begin{credits}

\subsubsection{\ackname} This study was supported by the Simons Foundation and the Breast Cancer Research Foundation (through grant MATH-23-001), and the NIH ROBIN cooperative group (grant U54CA274291). This work utilized resources from the High-Performance Computing Group at Memorial Sloan Kettering Cancer Center.

\subsubsection{\discintname}
The authors have no competing interests to declare that are relevant to the content of this article.

\end{credits}
%
%
%
%
\bibliographystyle{splncs04}
\bibliography{references2}
\end{document}